# Enhancing Small Dataset Classification Using Projected Quantum Kernels with Convolutional Neural Networks

A.M.A.S.D. Alagiyawanna
*Department of Computational Mathematics*
*University of Moratuwa*
Sri Lanka
alagiyawannaamasd.21@uom.lk

Asoka Karunananda
*Department of Computational Mathematics*
*University of Moratuwa*
Sri Lanka
asokakaru@uom.lk

A. Mahasinghe
*Department of Mathematics*
*University of Colombo*
Sri Lanka
anuradhamahasinghe@maths.cmb.ac.lk

Thushari Silva
*Department of Computational Mathematics*
*University of Moratuwa*
Sri Lanka
thusharip@uom.lk

*Abstract*—Convolutional Neural Networks (CNNs) have shown promising results in efficiency and accuracy in image classification. However, their efficacy often relies on large, labeled datasets, posing challenges for applications with limited data availability. Our research addresses these challenges by introducing an innovative approach that leverages projected quantum kernels (PQK) to enhance feature extraction for CNNs, specifically tailored for small datasets. Projected quantum kernels, derived from quantum computing principles, offer a promising avenue for capturing complex patterns and intricate data structures that traditional CNNs might miss. By incorporating these kernels into the feature extraction process, we improved the representational ability of CNNs. Our experiments demonstrated that, with 1000 training samples, the PQK-enhanced CNN achieved 95% accuracy on the MNIST dataset and 90% on the CIFAR-10 dataset, significantly outperforming the classical CNN, which achieved only 60% and 12% accuracy on the respective datasets. This research reveals the potential of quantum computing in overcoming data scarcity issues in machine learning and paves the way for future exploration of quantum-assisted neural networks, suggesting that projected quantum kernels can serve as a powerful approach for enhancing CNN-based classification in data-constrained environments.

## I. Introduction

Machine Learning (ML) is a part of computer science and AI that involves the design of data and algorithms to help artificial intelligence imitate the process of human learning, hence improving its accuracy over time. ML includes some main types of algorithms like Unsupervised, Supervised, and Reinforcement learning algorithms [1]. Various ML models use these algorithms. A machine learning model is an object (stored locally in a file) trained to recognize certain patterns. ML has evolved as Deep Learning with the rise of Artificial Neural Networks inspired by how our brain works. These models are better when the dataset becomes larger. Among deep learning algorithms, Convolutional Neural Networks have been popular graphical data [2].

Convolutional Neural Networks are smart computer models used for big classification tasks. They work well for image data, using layers to learn image patterns. This helps with recognizing images and videos. CNNs are very important for computer vision and image classification these days [3]. Convolutional Neural Networks help sort data by learning features through layers. They work well for sorting images because they handle the complexity of images. However, to use CNNs, we need lots of labeled data. The scarcity of data issue remains a significant challenge in Machine Learning. When datasets become smaller, even CNNs demand very low accuracy [4]. We hypothesize that leveraging Quantum Computing could potentially mitigate the ongoing issue of scarcity of data in ML solutions.

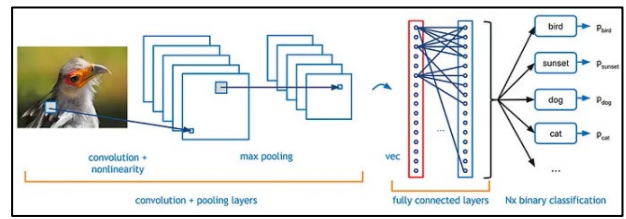

Fig. 1. Convolutional neural network (CNN) for classification [5].

Quantum Machine learning is a combination of quantum computing and machine learning. Instead of classical bits, quantum bits (widely known as 'qubits') are used in quantum computers. The classical bit has 0 and 1 while the qubit is in a superposition of two basis states $|0\rangle$ and $|1\rangle$ [6], which are called "Ket 0" and "Ket 1" in Dirac notation. In vector notation,

$$|0\rangle = \begin{bmatrix} 1 \\ 0 \end{bmatrix}, |1\rangle = \begin{bmatrix} 0 \\ 1 \end{bmatrix}$$

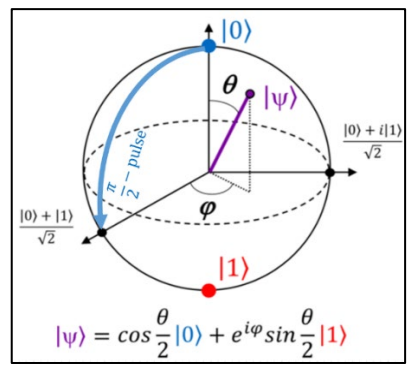

Fig. 2. Visualization of a qubit using bloch sphere representation [7].

It is noteworthy that the recent computing paradigms search for novel ways of incorporating non-determinism. In other words, making use of uncertainty for computations has been an intense topic of interest [8]. Quantum computing suggests leveraging quantum effects for this purpose. Due to the probabilistic nature of quantum measurement, using qubits enables us to use quantum effects such as superposition and entanglement. Superposition enables a system to exist in several states at a given time until it is measured [9], paving way to parallelization; and entanglement allows different types of protocols and algorithms that are inaccessible to classical machines [10]. This makes quantum computers exponentially faster that classical computers in doing certain tasks. Accordingly, quantum algorithms are designed, and complexity classes are also defined for quantum computers. This could detail very challenging interrelations of linked qubits, thus helping in the discovery of complex patterns more easily. Linking allows fast ways to look at many details— maybe tricks you normally parted your way to solve these problems quicker [11].

As classical circuits are made of classical gates, quantum circuits are made of basic quantum gates, that change the states of qubits. Accordingly, like AND gate, OR gate, NOT gate, etc. in classical circuits, quantum circuits also have specific basic gate sets such as CNOT gates, TOFOLLI gate and Hadamard gate [12]. However, their action is significantly different when compared with classical circuits. Quantum gates act as unitary operators on qubits and change their states in order to implement the specific steps given by quantum algorithms. It is noteworthy that the qubit has a useful visual representation, named the Bloch sphere, that helps to understand the role of the gates in changing its amplitudes.

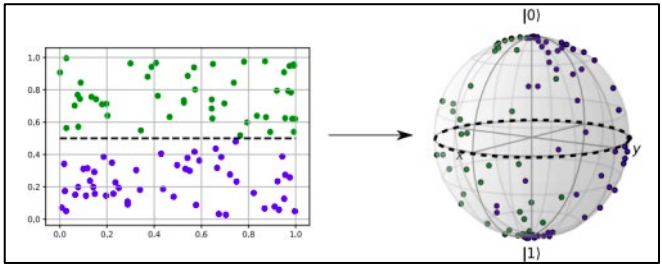

Fig. 3. A quantum feature map transforms the data points from, e.g., 2D input space to the bloch sphere [13].

An existing approach called Projected Quantum Kernel (PQK) as proposed by Huang et al. [14] can be adopted to improve image classification provided that data is scarce. The work of PQK is based on the modern technology of using quantum computers for data analysis and detecting subtle patterns through translating information into quantum states. However, the challenge is that these quantum states are not easily usable with conventional machine learning, thus PQK begins by performing a 'partial trace' to manage these quantum states and make them more amicable for use with the regular machine learning algorithms. This process aims at translating the read data back to the original classical form while maintaining some of the benefits of the quantum space. This balance lets PQK enhance feature extraction, and overall generalization; making it a useful tool that can generate great results with comparatively small datasets.

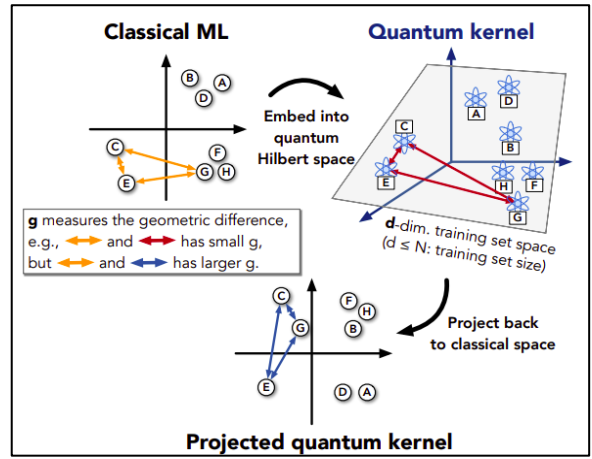

Fig. 4. A cartoon shows data points (A, B, ...) with arrows representing similarity measures (kernel functions) in classical and quantum ML models, highlighting the geometric difference (g) in these measures and the effective dimension (d) of the data in quantum space [14].

The upcoming sections of the paper are organized as follows. Section II describes the methodology of this research covering the Approach, Reducing dimensionality, Creating Quantum Circuits to extract features, Relabeling the data, Design of QCNN, and Experiments conducted to evaluate our approach to Quantum machine learning. Section III analyses and interprets the results generated by the experiments. Section IV presents the conclusion with a note on further work.

## II. METHODOLOGY

We hypothesize that leveraging Quantum Computing could potentially mitigate the compelling issue of scarcity of data in the construction of ML solutions. This section presents our Approach to quantum-based CNN and its design, implementation, and evaluation.

### A. Approach

To enhance accuracy and efficiency in ML solutions with limited data, our approach to integrating Quantum Projected Kernels with CNN employs 4 steps, namely, *reducing dimensionality, designing Qubits circuits, relabeling data,* and *designing QCNN.*

- Reducing the dimensionality: The dimensionality of raw data will be reduced as follows

$$Z = XW$$

where X is the input data, W is the projection matrix, and Z is the reduced-dimension data.

- Design of Qubit Circuit: Here Qubit Circuits are designed to extract features of the above reduced-dimensional data. High-dimensional quantum states are encoded in randomly selected rotation angles θ. The one-particle reduced density matrix (1RDM), denoted as $\rho^{(1)}$,

$$\rho^{(1)} = Tr_2[\rho]$$

where $Tr_2$ denotes the partial trace [15] over the degrees of freedom of particle 2.

- Relabeling data: according to extracted features using Projected Quantum Kernels.

- Designing a QCNN: with key layers that allow the network to process relabeled data.

Fig. 5. illustrates the steps in our approach to integrating CNN with Quantum computing.

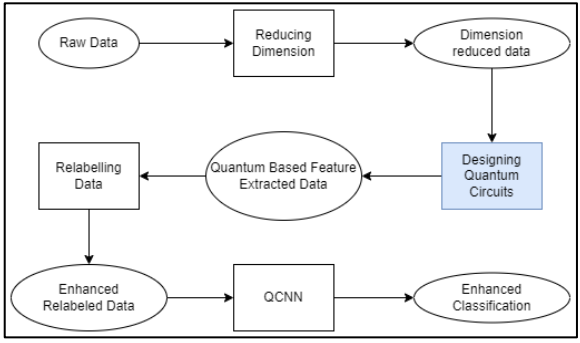

Fig. 5. Approach to integrate Quantum computing with CNN

Next, we briefly describe steps in our approach to quantum integration for CNN.

*B. Reducing Dimensionality*

Our research project under consideration took advantage of the two numbers in the MNIST dataset, namely 1 and 8 because they had different characteristics which made them perfect alternatives for binary classification purposes. Each grayscale picture of size 28x28 was converted in such a way that all pixels would take values between zero and one. This was achieved by normalizing every pixel to range from zero to one. For purposes of simplification and preparing it for use in quantum computation applications, we made use of Principal Component Analysis (PCA) [16] in compressing these pictures down to just ten main components. This dimensionality reduction retains the most significant features while making the dataset compatible with the constraints of current quantum computers.

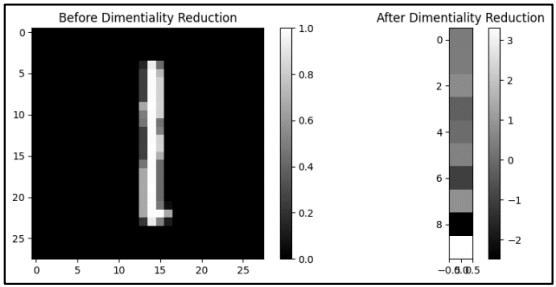

Fig. 6. Before and after dimensionality reduction of random data from MNIST data set.

*C. Quantum Circuit Design*

We begin by creating a wall of single qubit rotations [7], with each qubit in the circuit being rotated about the X, Y, and Z axis which are random angles that determine those rotations. As in Fig. 7., It is creating a wide variety specifically for the qubits at initialization. This guarantees that the input data is represented as a high-dimensional complex quantum state, leading to successful quantum feature extraction. Visualizing the circuit for a small number of qubits confirms the correct application of these random rotations.

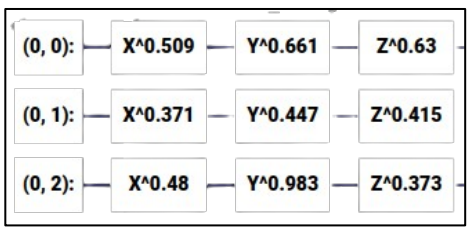

Fig. 7. Single qubit rotation circuit with 3 qubits.

Afterward, a Parametrized quantum circuit V(θ) [17] is created, wherein it should be noted that such a quantum circuit is composed of exponential that sum up Pauli operators [18] that interact pairwise on neighboring qubits (X, Y, Z). In this case, the parameters θ are used arbitrarily as symbols that determine the magnitudes of these operations. Therefore, it means that this specific circuit has the effect of converting the starting state into another state that represents classical input data in terms of quantum aspects and therefore makes it possible to resurface with PQK attributes. These features are then used to calculate the 1-RDM for each data point, providing a quantum-enhanced representation of the dataset.

The 1-RDM(One-Particle Reduced Density Matrix) [19] provides a simplified description of the quantum state by focusing on single-particle properties, effectively reducing the complexity of multi-particle systems. In our current project, the 1-RDM will be employed in isolating relevant characteristics of the modified quantum state. These features, derived from the expectation values of Pauli operators (X, Y, Z) on each qubit, offer a quantum-enhanced representation of the input data enabling more effective separation and analysis in quantum-classical hybrid algorithms.

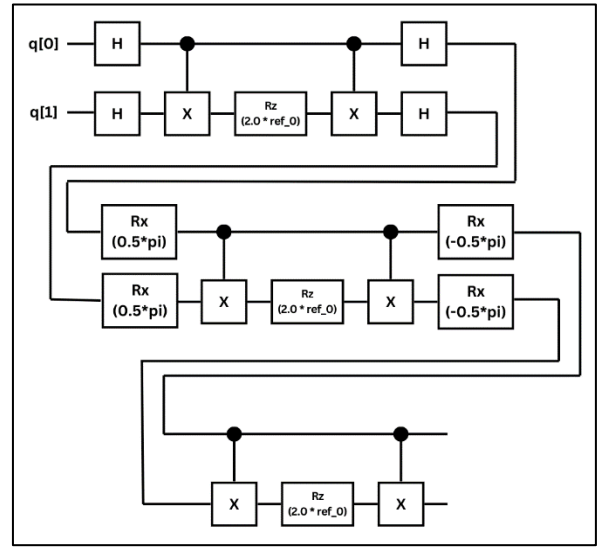

Fig. 8. Variational quantum circuit with only 2 qubits.

*D. Relabeling Data*

Relabeling with Projected Quantum Kernel features aims at re-calibrating the labels of the dataset to overemphasize the differentiation between quantum and classical models. This involves first calculating the kernel matrices concerning the PQK features and the original dataset. They encode the pairwise similarities in some high-dimensional feature space. The eigenvalues [20] and eigenvectors of such matrices convey information about the geometry of the structure and the separability of data transformed by quantum features. The aim is to maximize their geometric distance to the classical dataset kernels by applying a transformation that will optimally align eigenvectors [21]. Such eigenvectors are then used to generate new labels, putting more emphasis on quantum-specific traits inherent in the PQK features. This process is augmented with noise to ensure label robustness for generalizing separation between quantum and classical performance.

## E. Design of QCNN

The architecture used in this model is a sequential neural network whose convolutional and dense layers are specialized to process the input dimensions derived from PQK features. Convolutional layers are executed using ReLU [22] for non-linearity and trying to pick up complex patterns that exist in the data. The dense layers have sigmoid activation [23] functions that empower the output of their probabilistic predictions. The Binary Cross-Entropy loss function [24] measures the difference between the predicted and true labels. Besides, the Adam optimizer is utilized to enhance the efficiency of training in this model. The learning rate for this optimizer will be set at 0.003, which can deal effectively with gradient descent during the model training process. This architecture is optimized to ensure the best-in-class performance concerning classifying data, which is transformed by quantum-inspired features making up these layers for efficient pattern recognition and classification tasks.

## F. Evalution of QCNN

We used the MNIST[25] and CIFAR-10[26] datasets to conduct our experiments for training and testing our QCNN. The MNIST dataset contains 70,000 grayscale images of handwritten digits (0-9), with 60,000 for training and 10,000 for testing. The images are pre-processed to 28x28 pixel grids, normalized to squares no more than 20x20 pixels, and centered based on their weight mass. It is a widely used dataset for image recognition research. The CIFAR-10 dataset comprises 60,000 color images in 10 classes, including airplanes, automobiles, birds, cats, deer, dogs, frogs, horses, ships, and trucks. It has 50,000 training samples and 10,000 testing samples, each image sized at 32x32 pixels. This dataset is extensively used in machine learning and computer vision for benchmarking image classification algorithms and evaluating model performance due to its diversity and complexity.

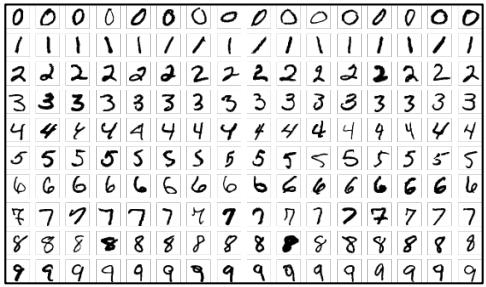

Fig. 9. The MINIST dataset [25].

First, the setup experiments preprocess the MNIST dataset to have normalized pixel values and filter it so that it covers only digits 1 and 8. Dimensionality is then reduced in this dataset to 10 components by applying Principal Component Analysis. With TensorFlow Quantum [27] and Cirq, the quantum feature extraction phase was performed upon the dimensionally reduced dataset to obtain Parametrized Quantum Kernels. These kernels are evaluated against classical features from the PCA-reduced dataset. It has models for a CNN-based architecture for PQK features and their classical CNN counterparts. Experiments range in dataset size from 100 to 1,000 samples, including model performance in terms of accuracy, training time, and scalability with 100 epochs and 32-sized batch sizes. The results are visualized with Matplotlib[28], showing Accuracy trends and training times relative to dataset size.

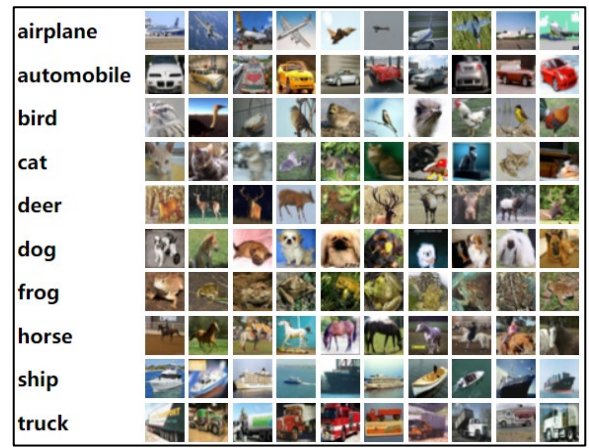

Fig. 10. CIFAR-10 dataset [26]

## III. RESULTS AND DISCUSSION

### A. Analysis of Accuracy

Fig. 11. and Fig. 12. shows the view of how accurate the PQK-extracted Convolutional Neural Network model and the Classical CNN model are as a function of dataset size for the MNIST and CIFAR-10 dataset respectively. Here, on the x-axis, we have the dataset size ranging from 100 to 1000 data points. Different y-axis values represent the accuracy from 0.0 to 1.0. We plot four lines for the training and testing accuracy for both models

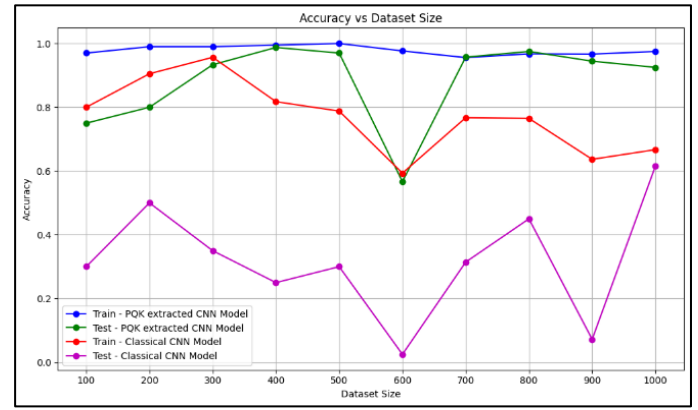

Fig. 11. Accuracy against the size of the MNIST dataset

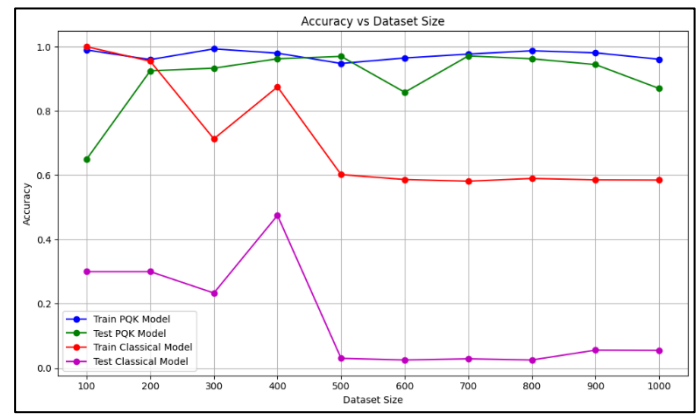

Fig. 12. Accuracy against the size of the CIFAR-10 dataset

For the CNN model with extracted PQK, the training accuracy is consistently high, close to 1.0, indicating that the model learns the training data almost perfectly. The test accuracy, independent of the model, is generally above 0.7 for the MNIST dataset and around 0.6 for the CIFAR-10 dataset.

Although the test accuracy drops noticeably with smaller dataset sizes—around 600 for MNIST and 500 for CIFAR-10 the overall generalization ability remains strong. In contrast, the Classical CNN model shows training accuracy ranging from 0.8 to 0.9 across both datasets, indicating good learning of the training data but with relatively poorer performance compared to the PQK model. The test accuracy for the Classical CNN model is significantly lower and more erratic, spanning from about 0.1 to 0.6 for the MNIST dataset and 0.1 to 0.5 for the CIFAR-10 dataset. This variability suggests poor generalization and a higher tendency for overfitting, especially with smaller datasets.

Compared to the classical CNN model, the PQK-extracted CNN model has higher results in both train and test accuracy. Together with high training accuracy, test accuracy is relatively stable across various dataset sizes suggests that the learning is effective and the generalization is robust. Testing accuracy for the PQK model varied but mostly was higher than the Classic CNN model. The result depicts the enhanced strength and stability in the projected quantum kernels. On the other hand, for the Classical CNN model, high variance in test accuracy indicates that this task either requires better regularization techniques or probably changes in architecture to improve its performance on unseen data.

*B. Analysis of Efficiency*

Fig. 13. and Fig. 14. illustrates the comparison of different training times of the PQK-extracted CNN model versus that of the Classical CNN model at different sizes of datasets for the MNIST and CIFAR-10 dataset respectively . The horizontal axis represents the size of the dataset, and the vertical axis represents the training time in seconds.

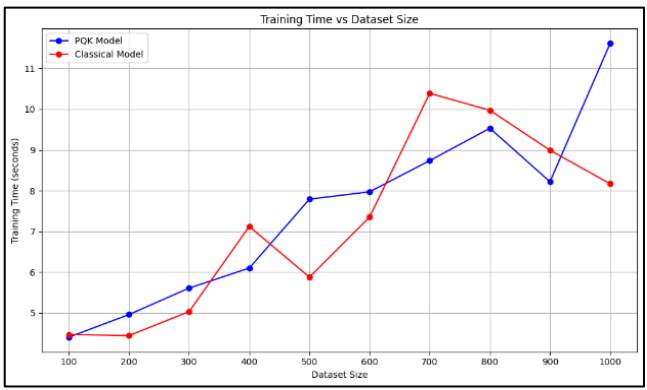

Fig. 13. Training time against the size of the MNIST dataset

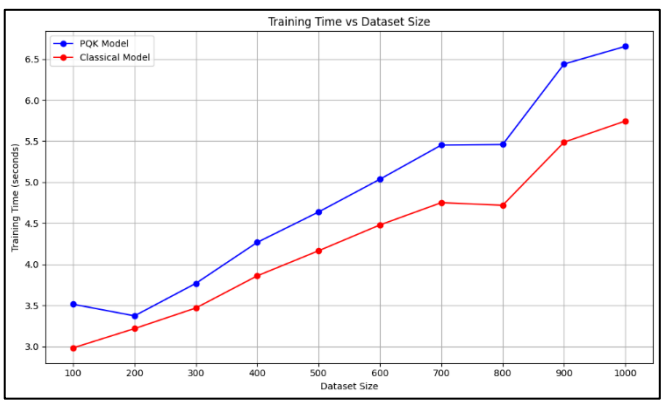

Fig. 14. Training time against the size of the CIFAR—10 dataset

The PQK model generally shows an increasing trend in training time as the dataset size increases. For the MNIST dataset, the training time peaks at 11 seconds with a dataset size of 1000, while for the CIFAR-10 dataset, it peaks at 6.5 seconds at the same dataset size. In contrast, the Classical CNN model also exhibits an increasing trend in training time with dataset size, but with notable fluctuations. For the MNIST dataset, the peak training time is around 10 seconds at a dataset size of 700 before stabilizing, and for the CIFAR-10 dataset, the peak training time is 5.5 seconds at a dataset size of 1000.

While both models increased the training time when the size of the datasets became larger, the PQK model typically had higher training times with larger datasets. This could be caused by additional computations needed during the extraction process for the projected quantum kernel. The Classical model expresses a more stable trend in training time with larger datasets compared to the PQK model and probably ensures much better scalability concerning training efficiency.

IV. CONCLUSION

Our research revealed that integrating Projected Quantum Kernels (PQK) into convolutional neural networks (CNNs) significantly improved feature extraction, proving highly beneficial for small datasets. Specifically, by using the PQK-enhanced CNN model, we conducted experiments on both the MNIST and CIFAR-10 datasets.

For the MNIST dataset, the PQK-enhanced CNN model demonstrated almost perfect training accuracy across all dataset sizes, indicating effective learning. The testing accuracy was generally above 70%, peaking at 95% with 1000 data samples, showcasing a robust generalization ability with minor fluctuations. This performance surpassed the Classical CNN model, which had a lower and more erratic test accuracy, peaking at around 60% with 1000 data samples, indicating a higher tendency to overfit. Similarly, for the CIFAR-10 dataset, the PQK-enhanced CNN model maintained high training accuracy and achieved a testing accuracy of around 60% with 1000 data samples, compared to the Classical CNN model's lower and more variable test accuracy, which peaked at around 50%.

While training times for the PQK-enhanced CNN were higher due to additional quantum kernel computations, reaching 11 seconds for 1000 data points on MNIST and 6.5 seconds on CIFAR-10, the classical CNN showed more stable and slightly lower times, peaking at 10 seconds for 700 data points on MNIST and 5.5 seconds for 1000 data points on CIFAR-10. These results suggest that while PQK-enhanced models offer significant accuracy gains, their scalability for larger datasets or real-time applications remains a challenge, highlighting the need to balance their benefits with computational efficiency for broader use cases.

The superior and more stable testing accuracy of the PQK model demonstrates the effectiveness of quantum-enhanced feature extraction in capturing complex patterns that traditional CNNs may miss. PQK serves as a crucial tool for developing robust models, especially in scenarios with limited data. This study shows that quantum computing principles can transcend some limitations of data availability, providing a boost to machine learning models.

Future work of our research aims at optimizing PQK extraction to enhance training speed and generalize its application to other neural networks and datasets, thus advancing the development of quantum-assisted neural networks in data-constrained environments. In doing so, we envisage developing a platform that accepts a given data set and builds the quantum gates to extract features enabling the training of a CNN and other Neural Networks. This platform will require users to have foundational knowledge in both quantum computing and neural networks, as these tools are designed with the expectation of expertise in these domains. Undoubtedly, this would be an excellent platform to experiment with Quantum machine learning solutions.